\documentclass[10pt,twocolumn,letterpaper]{article}

\usepackage{iccv}
\usepackage{times}
\usepackage{epsfig}
\usepackage{graphicx}
\usepackage{amsmath}
\usepackage{amssymb}

\usepackage{cuted}
\usepackage{caption}
\usepackage{booktabs}
\usepackage{makecell}
\usepackage{multirow}
\usepackage[cmyk]{xcolor}

\usepackage[pagebackref=true,breaklinks=true,letterpaper=true,colorlinks,bookmarks=false]{hyperref}

\iccvfinalcopy 

\def\iccvPaperID{5791} 

\ificcvfinal\fi

\begin{document}

\title{Narrator: Towards Natural Control of Human-Scene Interaction Generation\\ via Relationship Reasoning}

\author{
Haibiao Xuan$^1$ \hspace{0.5em} Xiongzheng Li$^1$ \hspace{0.5em} Jinsong Zhang$^1$ \hspace{0.5em} Hongwen Zhang$^2$ \hspace{0.5em} Yebin Liu$^2$ \hspace{0.5em} Kun Li$^{1}$ \\
$^1$Tianjin University \hspace{2em} $^2$ Tsinghua University\\
\vspace{4mm}
{\href{http://cic.tju.edu.cn/faculty/likun/projects/Narrator}{http://cic.tju.edu.cn/faculty/likun/projects/Narrator}}
}

\maketitle
\ificcvfinal\thispagestyle{empty}\fi

\begin{strip}
    \centering
    \vspace{-12mm}
    \includegraphics[width=1\textwidth]{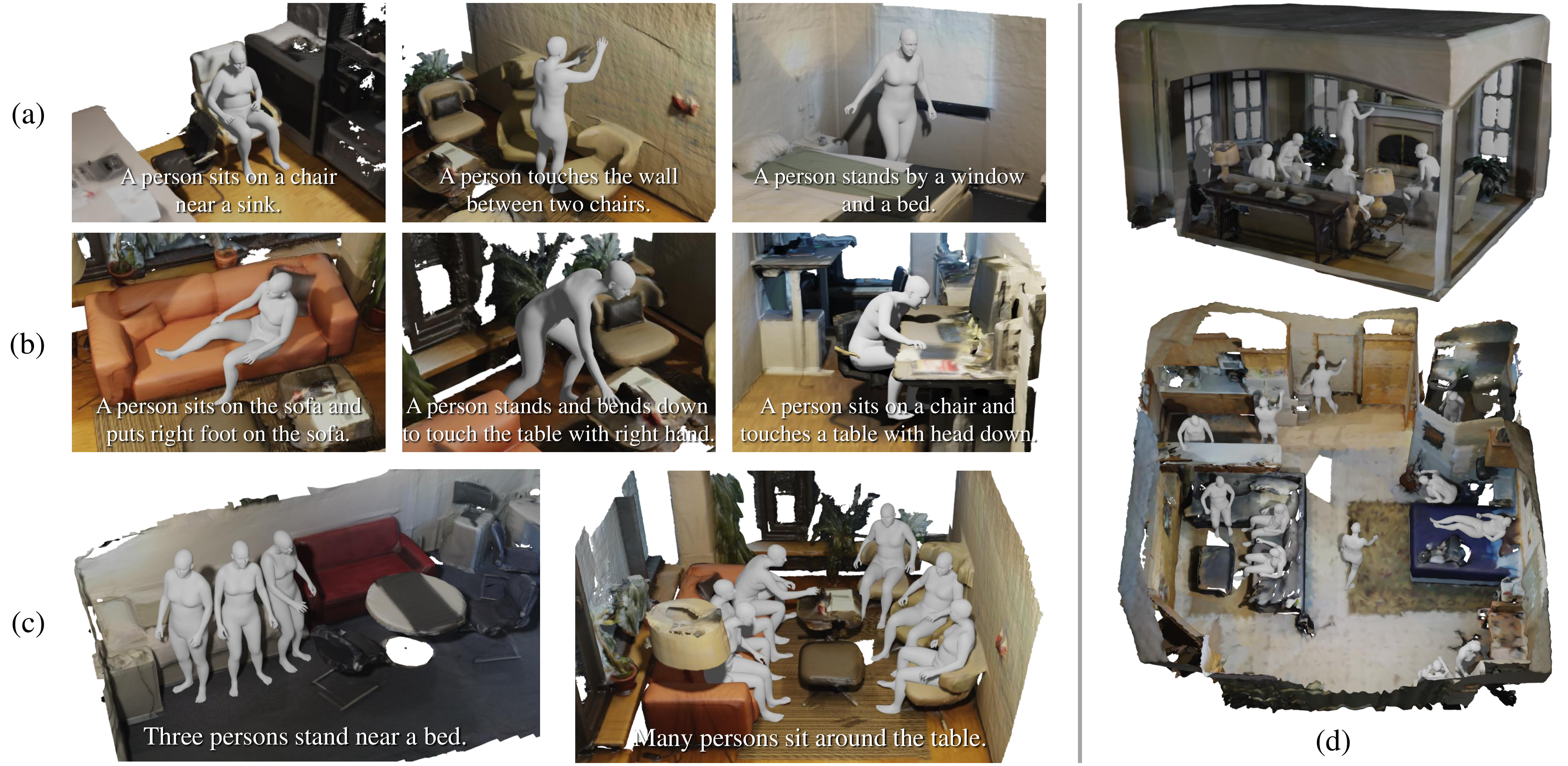}
    \vspace{-6mm}
    \captionsetup{type=figure}
    \caption{Given a textual description, our approach can naturally and controllably generate semantically consistent and physically plausible human-scene interactions for various cases: (a) interactions guided by spatial relationships, (b) interactions guided by multiple actions, (c) multi-human scene interactions, and (d) human-scene interactions combining the above interaction types, which cannot be generated using prior works.}
    \label{fig:teaser}
    \vspace{0mm}
\end{strip}

\begin{abstract}
   \vspace{-3mm}
   Naturally controllable human-scene interaction (HSI) generation has an important role in various fields, such as VR/AR content creation and human-centered AI. 
   However, existing methods are unnatural and unintuitive in their controllability, which heavily limits their application in practice. 
   Therefore, we focus on a challenging task of naturally and controllably generating realistic and diverse HSIs from textual descriptions. 
   From human cognition, the ideal generative model should correctly reason about spatial relationships and interactive actions. 
   To that end, we propose \textbf{Narrator}, a novel relationship reasoning-based generative approach using a conditional variation autoencoder for naturally controllable generation given a 3D scene and a textual description.
   Also, we model global and local spatial relationships in a 3D scene and a textual description respectively based on the scene graph, and introduce a part-level action mechanism to represent interactions as atomic body part states.
   In particular, benefiting from our relationship reasoning, we further propose a simple yet effective multi-human generation strategy, which is the first exploration for controllable multi-human scene interaction generation.
   Our extensive experiments and perceptual studies show that Narrator can controllably generate diverse interactions and significantly outperform existing works. 
   The code and dataset will be available for research purposes.
   
\end{abstract}

\newpage

\vspace{-2mm}
\section{Introduction}
\label{sec:Introduction}
\vspace{-1mm}
Throughout daily life, humans constantly interact with their surroundings and these interactions establish their relationships with the scenes.
Naturally controllable human-scene interaction (HSI) generation has significant value and numerous applications in areas such as VR/AR content creation, human-centered AI, and generating training data for other computer vision tasks.
In this paper, we tackle a challenging task of generating realistic and  plausible human-scene interactions from natural language textual descriptions, particularly exploring more liberal forms of HSIs with complex spatial relationships, multiple actions, and multiple persons, as shown in Fig. \ref{fig:teaser}.

Prior HSI methods \cite{zhang2020generating,hassan2021populating,wang2021synthesizing} mostly focus on the physical geometry between humans and scenes, but lacks the semantic control of generation.
Some works \cite{wang2022towards} further incorporate generative controls, but always coarsely describe them as action labels, not sentences.
A recent method, COINS \cite{zhao2022compositional}, specialises semantic control of interactions as combinations of actions and objects.
However, additional manual effort is required to explicitly specify object instances when faced with multiple objects of the same kind.
Moreover, binding actions to objects by force is not intuitive or reasonable.
For example, a natural case, ``standing by the window'', does not contain a direct and explicit interaction object, and COINS cannot deal with it.
These unnatural and constrained control ways fall short of meeting the needs of users and limit their applicability.

Humans usually naturally describe people who have diverse interactions in different places through spatial perception and action recognition.
Thus, an ideal generative model should correctly reason about \textit{spatial relationships} to obtain the human position that respects textual descriptions while exploring degrees of freedom about \textit{interactive actions} to generate natural interactions. 
Specifically, \textit{spatial relationships} can be represented as the interrelationship among different objects in a scene or a local area, and \textit{interactive actions} are specified by atomic body part states, such as a person's left feet treading, torso leaning, right hand tapping, and head bowing.
How to reason about these relationships and utilize these powerful cues for naturally controllable generation is a pressing problem. 

To address these issues, we propose \textit{Narrator}, a novel generative approach that incorporates a transformer-based conditional variational auto-encoder (cVAE) framework and leverages relationship reasoning to naturally produce diverse and plausible HSIs given the scene and textual description.
The diversity and complex interrelationship of objects in scenes can lead to misjudgements of human position and unnatural interactions. 
Therefore, instead of understanding scenes or specific objects in isolation as previous works, we employ the scene graph to represent spatial relationships and propose a Joint Global and Local Scene Graph (JGLSG) mechanism to provide global perception for subsequent localization, allowing for interaction generations guided by spatial relationships (in Fig. \ref{fig:teaser} (a)). 
As body part states are key for modeling realistic and text-faithful interactions, we introduce a Part-Level Action (PLA) mechanism to establish the correspondence between human body parts and actions, allowing for interaction generations guided by multiple actions (in Fig. \ref{fig:teaser} (b)).
Ultimately, we feed the multi-modal features extracted by JGLSG, PLA and PointNet++ \cite{qi2017pointnet++} as a joint conditional embedding into cVAE, thus obtaining a unified latent space of the human body.
To train and evaluate our method, we annotate multi-level text descriptions from coarse to fine for each frame of the PROX dataset \cite{hassan2019resolving}.

In real-world scenes, there are more situations where multiple people are interacting independently or in a connected way.
Unfortunately, there is no work that solves this problem in an automatic and controlled way, but rather requires certain expertise and manual effort \cite{li2019putting, zhang2020generating}.
Also, a straightforward way by using a single-person method like COINS \cite{zhao2022compositional}, \textit{i.e.}, sequential per-person generation and optimization, does not properly understand multi-person text descriptions, leading to unreasonable spatial distributions and unnatural interactions of the generated results.
In contrast, benefiting from the flexibility and reusability of our JGLSG and PLA mechanisms, we propose a simple yet effective multi-human generation strategy.
We reason out each person's interaction information from the text and globally update each generation to establish their relationships, thus achieving a better spatial distribution than simple multiple generation.
To our knowledge, this is the first naturally controllable and user-friendly generative model for multi-human scene interaction (MHSI) (in Fig. \ref{fig:teaser} (c)).

In brief, our contributions can be summarized as:
\vspace{-2.5mm}
\begin{itemize}
\setlength{\itemsep}{-1.2mm}
\item we present \textit{Narrator}, a new generative method for naturally controllable human scene interaction generation given textual descriptions in natural language.
\item we propose the JGLSG and PLA mechanisms for relationship reasoning considering narrator's perspective.
\item we propose the first naturally controllable MHSI generation strategy to approximate the real world.
\end{itemize}


\vspace{-3mm}
\section{Related work}
\label{sec:Related_work}
\vspace{-2mm}

\textbf{Human-Scene Interaction (HSI).}
Human-scene interaction, a challenging task in computer vision, recently has received increasing attention.
Early HSI methods \cite{kim2014shape2pose, savva2016pigraphs, tan2018and, li2019putting, chen2019holistic++, monszpart2019imapper} focused on scene affordance and function understanding, but the lack of relevant high-quality datasets and valid human representations lead to low-fidelity interaction and poor results.
PiGraphs \cite{savva2016pigraphs} learns the probability distribution of each {verb-object} category from real-world interactions to generate interaction snapshots, but the simple representation of the 3D human as a skeleton prevents reasoning about contact details.
Aided by the parametric body model SMPL-X \cite{pavlakos2019expressive} and the dataset PROX \cite{hassan2019resolving} recording human activities in 3D scenes, efforts in recent years continually iterate and refine towards realism and naturalism.
Zhang \textit{et al.} \cite{zhang2020generating} trains a conditional variation autoencoder (cVAE) to predict semantically plausible 3D human poses with scene depth and semantics, and apply geometric constraints for physical plausibility.
POSA \cite{hassan2021populating} proposes a human-centric contact map that encodes contact probability and semantic information for each vertex on the body mesh, and uses these to guide the search for its most likely position in the scene.
Although these methods \cite{zhang2020generating, zhang2020place, hassan2021populating, huang2023diffusion} model interactions with different representations, they cannot support controllable interaction synthesis.
Given the action sequence, Wang \textit{et al.} \cite{wang2022towards} propose a three-stage framework to place humans into scenes, produce feasible paths, and complete motion synthesis. 
COINS \cite{zhao2022compositional} represents semantic control as combinations of actions and objects, similar to PiGraph \cite{savva2016pigraphs}, and combine atomic interactions into compositional interactions. 

COINS is the SOTA HSI method, and the most relevant work, but our work has the superiority in many aspects:
1) The control way of COINS is not intuitive and requires additional manual selections for object instances, while our method is fully automatic conditioned on natural language description;
2) Our method has more flexible spatial localization ability and can handle non-direct interactions (\textit{e.g.}, standing by the window), while COINS fails;
3) COINS has limited interaction types and combinations, whereas ours is more diverse and we can simultaneously support more constraints (\textit{e.g.}, left/right hand lift, bend and crouch);
4) We are the first to explore and achieve controllable MHSI generations via our relationship reasoning.

\textbf{Text-guided Action \& Object Grounding.}
Grounding human actions and objects in scenes from textual descriptions are important and meaningful tasks that have received much exploration. 
For text-guided action grounding, recent works explore advances in natural language with many amazing results \cite{ahn2018text2action, ahuja2019language2pose, youwang2022clip, song2022actformer, hong2022avatarclip, ghosh2021synthesis, petrovich2022temos, athanasiou2022teach, guo2022generating}. 
CLIP-Actor \cite{youwang2022clip} utilizes multi-modal perception and semantic matching to synthesize the best matching action sequences from a text-visual coupling perspective.
Guo \textit{et al.} \cite{guo2022generating} propose a two-stage pipeline to implement the prediction from input text to visual action length and then to motion generation. 
On the other hand, 3D object grounding aims to locate the most relevant target object in 3D point cloud scenes conditioned on textual descriptions \cite{chen2020scanrefer, yang2021sat, feng2021free, roh2022languagerefer, luo20223d}. 

Different from the above-mentioned, our approach takes more account of possible human interactions in the scene and refines these into the body part states. Meanwhile, for better localization and grounding, we unite position features encoded from textual descriptions and 3D scenes into the conditional embedding of the cVAE.

\vspace{-3mm}
\section{Overview}
\label{sec:Overview}
\vspace{-2mm}

\begin{figure*}
  \centering
  \vspace{-4mm}
  \includegraphics[width=1.0\linewidth]{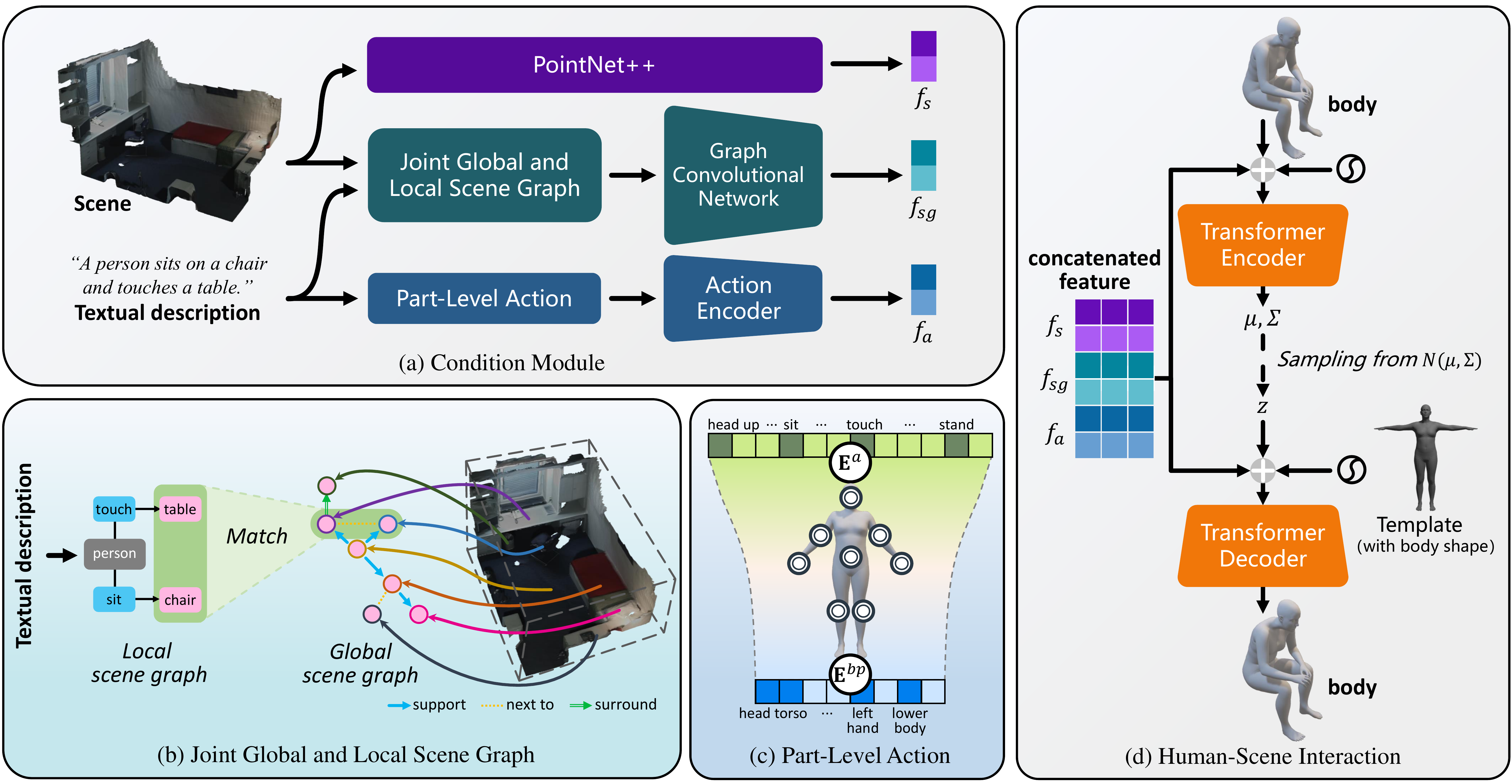}
  \vspace{-7.5mm}
  \caption{Overview of the proposed \textit{Narrator} framework. Given a scene and a textual description, multi-modal features including scene features, scene graph features, and action features are extracted (a), where the latter two are reasoned through our Joint Global and Local Scene Graph (b) and Part-Level Action (c), respectively. These features are then concatenated as a joint conditional embedding and fed into the transformer-based cVAE framework for human-scene interaction (d).}
  \vspace{-6mm}
  \label{fig:pipeline}
\end{figure*}

Our goal is to naturally generate human-scene interactions that are semantically consistent with textual descriptions and physically plausible with scenes. 
Fig. \ref{fig:pipeline} shows the framework of our approach. 
To this end, we propose a novel generative approach, \textit{Narrator}, with a transformer-based conditional Variational Auto-Encoder (cVAE) network architecture (Sec. \ref{sec:Network}). 
Specifically, in contrast to existing works that consider scenes or objects in isolation, we design a Joint Global and Local Scene Graph (JGLSG) mechanism to reason about complex spatial relationships for global localization perception (Sec. \ref{sec:JGLSG}). 
In addition, people simultaneously engage in diverse interaction activities with different body parts. This inspired us to introduce a Part-Level Action (PLA) mechanism for realistic and diverse interactions (Sec. \ref{sec:PLA}).
Meanwhile, we introduce an Interaction Bisector Surface (IBS) loss to obtain better generation results during scene-aware optimization (Sec. \ref{sec:Optimization}). 
We further broaden into multi-human fields and ultimately facilitate the first step to MHSI (Sec. \ref{sec:MHSI}). 

Here we give the representation of the scene, the body mesh and the textual description. 
We denote the scene as $S=\left(V_s, S_s\right)$, where $V_s$ and $S_s$ stand for vertices and per-vertex semantic labels, respectively. 
We represent the 3D human body mesh using a SMPL-X model \cite{pavlakos2019expressive} and a POSA representation \cite{hassan2021populating}. 
Specifically, the SMPL-X body mesh $M_{\text{SMPL-X}}=(V_b, F_b)$ with vertices $V_b \in \mathbb{R}^{10475 \times 3}$ and triangles $F_b$, is parameterized by a differentiable function $F(t, r, \beta, p, h)$, where $t \in \mathbb{R}^3$ is the global translation, $r \in \mathbb{R}^6$ is a continuous representation \cite{zhou2019continuity} of the global orientation, $\beta \in \mathbb{R}^{10}$ is the body shape parameters, $p \in \mathbb{R}^{63}$ is the body pose parameters, and $h \in \mathbb{R}^{24}$ is the hand pose parameters. We also extract contact labels $L_b$ using POSA. Overall, we define the body mesh as $M = (V_b, F_b, L_b)$.
Besides, the textual description about HSI includes various levels of interaction detail, defined as a sequence of words $W_{1: N}=[w_1, \ldots, w_N]$ from the English vocabulary. 

\vspace{-1mm}
\section{Method}
\label{sec:Method}
\vspace{-1mm}

\subsection{Network Architecture}
\label{sec:Network}
\vspace{-1mm}
For naturally controllable HSI generation, we employ a transformer-based cVAE architecture that can handle multi-modal information including scenes and textual descriptions, and model the probability $p\left(M \mid S, W_{1:N}\right)$. We describe the details of each part as follows.

\noindent\textbf{Condition Module.}
The condition module takes a 3D scene and a textual description as input, and outputs a joint conditional embedding. 
First, we employ PointNet++ \cite{qi2017pointnet++} to extract the scene $S$ into 256-dimension scene features $f_{s}$. 
Then, to reason the spatial and structural relationships, the scene and textual description are simultaneously input to the JGLSG to obtain the scene graph including human nodes, and the scene graph feature $f_{sg}$ is obtained by encoding it using a Graph Convolutional Network \cite{li2019deepgcns}. 
In addition, to reason about the human-action relationship, the action combination parsed from the textual description is fed into the PLA for mapping to atomic body part states, and then the Action Encoder encodes them as the action feature $f_a$. 
Finally, these three features are concatenated as a joint conditional embedding $f_{ce}$. 

\noindent\textbf{Transformer Encoder.}
We first utilize a fully-connected layer (FC) to encode human body mesh $M$ as a high-level embedding and concatenate it with joint conditional embedding $f_{ce}$ as input. 
On top of the encoder, we apply average pooling to the output, followed by another FC to predict the Gaussian distribution $\mathcal{Q}\left(z \mid S, W_{1:N}, M\right)$. 
Finally, we sample the latent code $z$ from the distribution using the reparameterization trick, as one of the decoder inputs. 

\noindent\textbf{Transformer Decoder.}
In the decoder, we use a SMPL-X template body with person-dependent shape parameters as the body token to improve the generalization of our model and achieve finer-grained control. 
We concatenate the latent code $z$ with the joint conditional embedding $f_{ce}$ as another input for the decoder. The output body mesh is fed into the SMPL-X regressor \cite{zhao2022compositional} to regress with consistent SMPL-X body parameters for loss supervision. 

\vspace{-1mm}
\subsection{Joint Global and Local Scene Graph}
\label{sec:JGLSG}
\vspace{-1mm}
Reasoning about spatial relationships can provide scene-specific clues to the model, which plays an important role in achieving natural controllability for HSI. 
Therefore, we design a Joint Global and Local Scene Graph mechanism, which is implemented through the following three steps.

\vspace{-0.3mm}
\noindent\textbf{Global Scene Graph Generation:} Given the scene, we use a model \cite{zhou2019scenegraphnet} pre-trained on 3DSGG \cite{wald2020learning} to generate a global scene graph, \textit{i.e.}, 
$\mathcal{GSG}=(\mathcal{V}, \mathcal{E}), \mathcal{V}=\left\{o_i\right\}_{i=1}^n, \mathcal{E}=\{(o_i, r_{ij}, o_j)_k\}_{k=1}^m$,
where $o_i, o_j$ are the objects with category labels, $r_{i j}$ is the relationship between $o_i$ and $o_j, n$ is the number of objects, and $m$ is the number of relationships.

\vspace{-0.3mm}
\noindent\textbf{Local Scene Graph Generation:} Our model adopts an off-the-shelf semantic parsing toolkit \cite{wu2019unified} to recognise the syntactic structure of the textual description and extract a set of triplets $\{T_{ij}\}$, where $T_{ij} = (s_i, p_{ij}, o_j)$ defines a triplet of subject-predicate-object.
The output of the syntactic parser is not sufficient to represent the human spatial location, especially the number of objects in the scene.
Hence, we build quantity checker for detecting its quantifier expression, and duplicating object nodes for the local scene graph $\mathcal{LSG}$ (\textit{e.g.}, ``three plants'' in textual descriptions → three ``plant'' nodes in $\mathcal{LSG}$).

\vspace{-0.3mm}
\noindent\textbf{Scene Graph Matching:}
Then, we correspond the local scene graph to the nodes in the global scene graph based on same object semantic labels.
During this process, two object category concepts can be matched if there is an overlap between their synsets, lemmas, or hypernyms (\textit{e.g.}, ``armchair'' → ``chair'').
According to the corresponding result of each object, we add a virtual human node by extending the edge relationships for providing the generated position, so we obtain a final scene graph $\mathcal{SG}$ that is consistent with both the scene and the textual description.

\subsection{Part-Level Action}
\label{sec:PLA}

Human interactions in the scene are composed of atomic body part states, and hence we propose a part-level action mechanism to select the important parts and disregard the irrelevant parts from the given interactions. 
Specifically, we explore richer interactive actions than existing works \cite{wang2022towards, zhao2022compositional} and correspond these possible actions to the five main human body parts: head, torso, left/right arm, left/right hand, and left/right lower body, as shown in Tab. \ref{tab:PLA}. Also, we use the one hot vectors $E^{a}$ and $E^{bp}$ to represent these actions and body parts, respectively. 
Then we concatenate them based on our proposed correspondence for subsequent encoding. 

For interaction generation guided by multiple actions, the attention mechanism of the transformer network is employed to learn the different part states of the SMPL-X body mesh. 
Given a combination of interactive actions, the attention between its corresponding body part and all other actions for each action, is automatically masked for each action. 
Taking the example of ``a person crouches on the floor using a cabinet'', crouching corresponds to the state of the lower body, and hence the attention of other parts tokens will be masked to zero.

\begin{table}
  \centering
  \begin{tabular}{@{}p{1.7cm}p{6cm}@{}}
    \toprule
    Part & Interaction \\
    \midrule
    Head & head up, head down, head left, head right \\
    Torso & sit, sit down, lean, lie, lie down \\
    (L/R) Arm & stretch, bend, straight, supported, raise, put \\
    (L/R) Hand & touch, use, hold, support, supported, type, write, open \\
    (L/R) Lower   body & stand, stand up, step, step up, step down, step back, walk, run, move, crouch, turn around, raise, put \\
    \bottomrule
  \end{tabular}
  \vspace{-3mm}
  \caption{List of body part actions, where (L/R) indicates the left and right of the part. }
  \vspace{-5mm}
  \label{tab:PLA}
\end{table}

\vspace{-2mm}
\section{Scene-aware Optimization}
\label{sec:Optimization}
\vspace{-2mm}
We perform scene-aware optimization with geometric and physics constraints to improve the generation results, following \cite{zhang2020generating, hassan2021populating, zhao2022compositional}.
Throughout the optimization process, it ensures that the generated poses do not deviate while encouraging contact with the scene and constraining the body mesh to avoid interpenetration with the scene surface.
Given the scene mesh $S$ and generated SMPL-X parameters, the optimization loss is given by:
\begin{equation}
  \mathcal{L}_{opt} =\mathcal{L}_\text{cont}+\mathcal{L}_\text{coll}+\mathcal{L}_\text{IBS}+\mathcal{L}_\text{reg}, 
  \label{eq:opt}
\end{equation}
where $\mathcal{L}_{cont}$ encourages body vertices to contact with the scene mesh, $\mathcal{L}_{coll}$ is the signed-distance-based collision term defined in \cite{zhang2020generating}, and the $\mathcal{L}_{reg}$ is a regularizer that penalizes SMPL-X parameters deviating from the initialization. 

A further addition we make over existing HSI methods is adopting the Interaction Bisector Surface (IBS) \cite{zhao2014indexing}, which is the set of points equidistant from two sets of points sampled on the scene and the human, respectively. For our task, we modify it as additional loss supervision $\mathcal{L}_\text{IBS}$:
\setlength\abovedisplayskip{0.12cm}
\setlength\belowdisplayskip{0.10cm}
\begin{equation}
  \mathcal{L}_\text{IBS}=\sum_{v^p \in V} d^p_s,
  \label{eq:ibs}
\end{equation}
where $V$ denotes the set of all points in the IBS point set that satisfies either penetration or corresponds to the body vertices with contact labels, and $d^p_s$ indicates the distance from point $v^p$ to the scene.

For more details regarding the training and optimization, please refer to the Supp.Mat.

\section{Multi-Human Scene Interaction}
\label{sec:MHSI}
\vspace{-2mm}
In real-world scenes, many situations are not just one person interacting with the scene, but multiple people interacting independently or in an associated way.
However, due to the lack of MHSI datasets, existing methods fail to handle this task in a controlled and automatic manner, but require additional manual effort.
To this end, we propose a simple but effective strategy for MHSI, using only existing single human datasets.

Given a textual description about MHSIs, our model first parses it into multiple local scene graphs $\mathcal{LSG}_i$ and human interactive actions $\mathcal{IA}_i$. We define the candidate set as $\mathcal{S}_c=\{(\mathcal{LSG}_i, \mathcal{IA}_i)\}^{l}_{i=1}$, where $l$ is the number of people. For each element of the candidate set $\mathcal{S}_c$, we first feed it into \textit{Narrator} together with the scene $\mathcal{S}$ and the corresponding global scene graph $\mathcal{GSG}$, subsequently performing the optimization process. To handle collisions between humans, we additionally introduce a loss $\mathcal{L}_\text{HH}$ during the optimization process as follows:
\setlength\abovedisplayskip{0.15cm}
\setlength\belowdisplayskip{0.10cm}
\begin{equation}
  \mathcal{L}_\text{HH} = \sum_{i} \Psi_\text{H}(v_i),
\end{equation}
where $\Psi_\text{H}(v_i)$ denotes the signed distance of vertex $v_i$ of the generated body mesh to other persons.

Then, when the optimization loss is below a threshold determined by experimental experience, we accept this generation and simultaneously update $\mathcal{GSG}$ by adding the human node. Otherwise, we consider the generation result implausible and update $\mathcal{GSG}$ by pruning the corresponding object node.
It is worth noting that this update way establishes the relationship between each generation and the previous results and avoids a certain degree of crowding, allowing for a better spatial distribution and more realistic interaction than simple multiple generation.

These procedures can be formulated as:
\setlength\abovedisplayskip{0.12cm}
\setlength\belowdisplayskip{0.12cm}
\begin{equation}
\begin{aligned}
  &I_g= \text{Narrator}(S, \mathcal{GSG}_i, \mathcal{LSG}_i, \mathcal{IA}_i),\\
  &I_g = \text{Opt}(I_g),\\
  &\mathcal{GSG}_{i+1} = \text{Update}(I_g).
  \label{eq:MHSI}
\end{aligned}
\end{equation}

With this design, our model can deal with multi-person interaction generation trained on existing datasets.

\section{Experiments}
\label{sec:Experiments}
\vspace{-1mm}
\subsection{Datasets}
\label{sec:Datasets}
\vspace{-1mm}
As there is no existing dataset suitable for our task, we annotate all video frames from PROX \cite{hassan2019resolving} with their corresponding textual descriptions about interactions to evaluate our natural and controllable HSI generation. 
Thus, similar to Sr3D \cite{achlioptas2020referit3d}, we design a combinatorial template to accurately describe human interaction and location in natural language descriptions:
\setlength\abovedisplayskip{0.1cm}
\setlength\belowdisplayskip{0.1cm}
\begin{equation}
\begin{aligned}
    &<\text{subject}> <\text{action}><\text{object-class}>\\&<\text{spatial-relationship}><\text{anchor-class(es)}>,
\end{aligned}
\end{equation}
where the action is taken from the motion labels in BABEL \cite{punnakkal2021babel} and can be described as multiple actions.
Please refer to the Supp.Mat. for details.

To demonstrate the generalisation capability of the proposed framework, we further evaluate it on Matterport3D \cite{chang2017matterport3d} and ScanNet \cite{dai2017scannet}, which both provide large-scale reconstructed 3D scenes. Please note that our framework does not utilize Matterport3D and ScanNet for training.

\vspace{-1.5mm}
\subsection{Baselines}
\label{sec:Baselines}
\vspace{-1mm}
Currently available methods do not allow for naturally controllable HSI generations directly from textual descriptions. Thus, we modify three state-of-the-art HSI methods and train their official models using the same dataset. 

\noindent\textbf{PiGraph-Text.}
PiGraph \cite{savva2016pigraphs} generates scene placement and human skeletons from interaction category specifications. We remove the scene placement step, represent the body with SMPL-X model and replace verb-noun pairs with textual descriptions. We denote this modified PiGraph variant as PiGraph-Text. 

\noindent\textbf{POSA-Text.}
POSA \cite{hassan2021populating} populates 3D scenes with humans guided by per-vertex contact features, but lacks effective control. To incorporate semantic guidance, we first generate body meshes that match textual descriptions and then place them in the appropriate positions using POSA. We denote this modified POSA variant as POSA-Text. 

\noindent\textbf{COINS-Text.}
COINS \cite{zhao2022compositional} synthesizes HSIs given interaction semantics as action-object pairs. However, COINS only works for relatively limited interaction combinations and cannot handle complex spatial relationships. Therefore, we extend more interactions and modify it to a two-stage process: first find the area of possible interactions based on BERT and then run COINS as is within that area. We denote this modified COINS variant as COINS-Text and present fairness experiments for COINS in the Supp.Mat.

\begin{figure}
  \centering
  \vspace{-2mm}
  \includegraphics[width=1.0\linewidth]{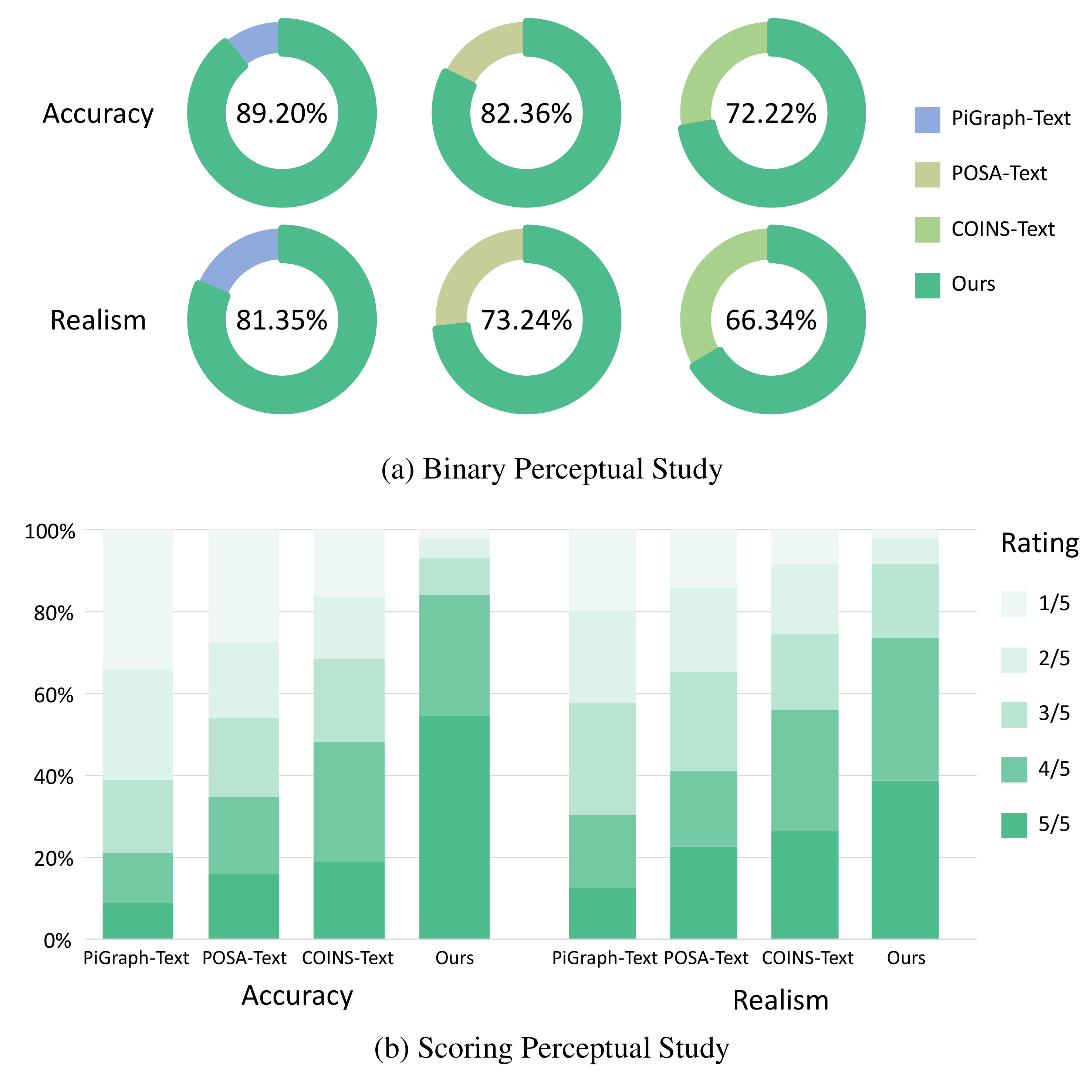}
  \vspace{-9mm}
  \caption{Perceptual study results (anonymized and order-randomized) comparing our approach against three baselines. In the binary perception study (a), the percentage numbers indicate the proportion of respondents who preferred our approach compared to another baseline. In the scoring perception study (b), the different colored bars indicate the percentage of corresponding ratings.}
  \vspace{-6mm}
  \label{fig:results1}
\end{figure}

\vspace{-1.5mm}
\subsection{Evaluation Metrics}
\label{sec:Metrics}
\vspace{-1mm}
\noindent\textbf{Physical Plausibility.} From the physical perspective, we evaluate the contact and non-collision scores between the generated body and scene mesh. The former is calculated as the proportion of actual contact vertices for all body vertices with contact labels, while the latter is calculated as the ratio of the number of body vertices with non-negative scene SDF values and the number of all body vertices.

\noindent\textbf{Diversity.} We perform K-Means (K = 50) clustering on the generated human-scene interactions and report: (1) the entropy of the cluster ID histogram, and (2) the cluster size, \textit{i.e.}, the average distance between the cluster center and the samples belonging in it.

\noindent\textbf{Perceptual Study.}
We evaluate the interaction realism and the accuracy of generated results by conducting perceptual studies, which consisted of two main parts: 
(1) we perform a binary-choice perceptual study in which samples generated by different methods based on the same textual description are displayed, and the respondents are asked to select the more realistic and natural sample; 
(2) the respondents are also asked to rate the accuracy and the consistency between shown interaction samples and textual descriptions, from 1 (strongly disagree) to 5 (strongly agree). 
Please note that the order is randomly swapped in each display. 
For more details regarding the perceptual studies, please refer to the Supp.Mat.

\begin{table*}
  \centering
  \setlength{\tabcolsep}{3mm}
  \begin{tabular}{@{}cccccc@{}}
    \toprule
    \multirow{2}{*}{Methods} & \multirow{2}{*}{Perceptual Accuracy ($\uparrow$)} & \multicolumn{2}{c}{Physical Plausibility} & \multicolumn{2}{c}{Diversity} \\ \cmidrule(l){3-6}
    & & Contact ($\uparrow$) & Non-Collision ($\uparrow$) & Entropy ($\uparrow$) & Cluster Size ($\downarrow$) \\ \midrule
    PiGraph-Text & 2.81$\pm$1.30 & 0.84 & 0.81 & 3.56 & 1.75  \\
    POSA-Text & 3.14$\pm$1.43 & 0.72 & 0.96 & 3.73 & 1.96  \\
    COINS-Text & 3.48$\pm$1.35 & 0.91 & 0.93 & 3.98 & 1.83  \\
    Ours & \textbf{4.02}$\pm$\textbf{0.97} & \textbf{0.94} & \textbf{0.98} & \textbf{4.16} & \textbf{1.54} \\
    \bottomrule
  \end{tabular}
  \vspace{-1.5mm}
  \caption{Quantitative comparison with three baselines. Perceptual accuracy is used to evaluate the degree of consistency with textual descriptions. Contact score and non-collision score are used to evaluate interaction realism and plausibility. Entropy and cluster size are used to evaluate interaction diversity.} 
  \label{tab:DB}
  \vspace{-4mm}
\end{table*}

\begin{figure*}
  \centering
  \includegraphics[width=1\linewidth]{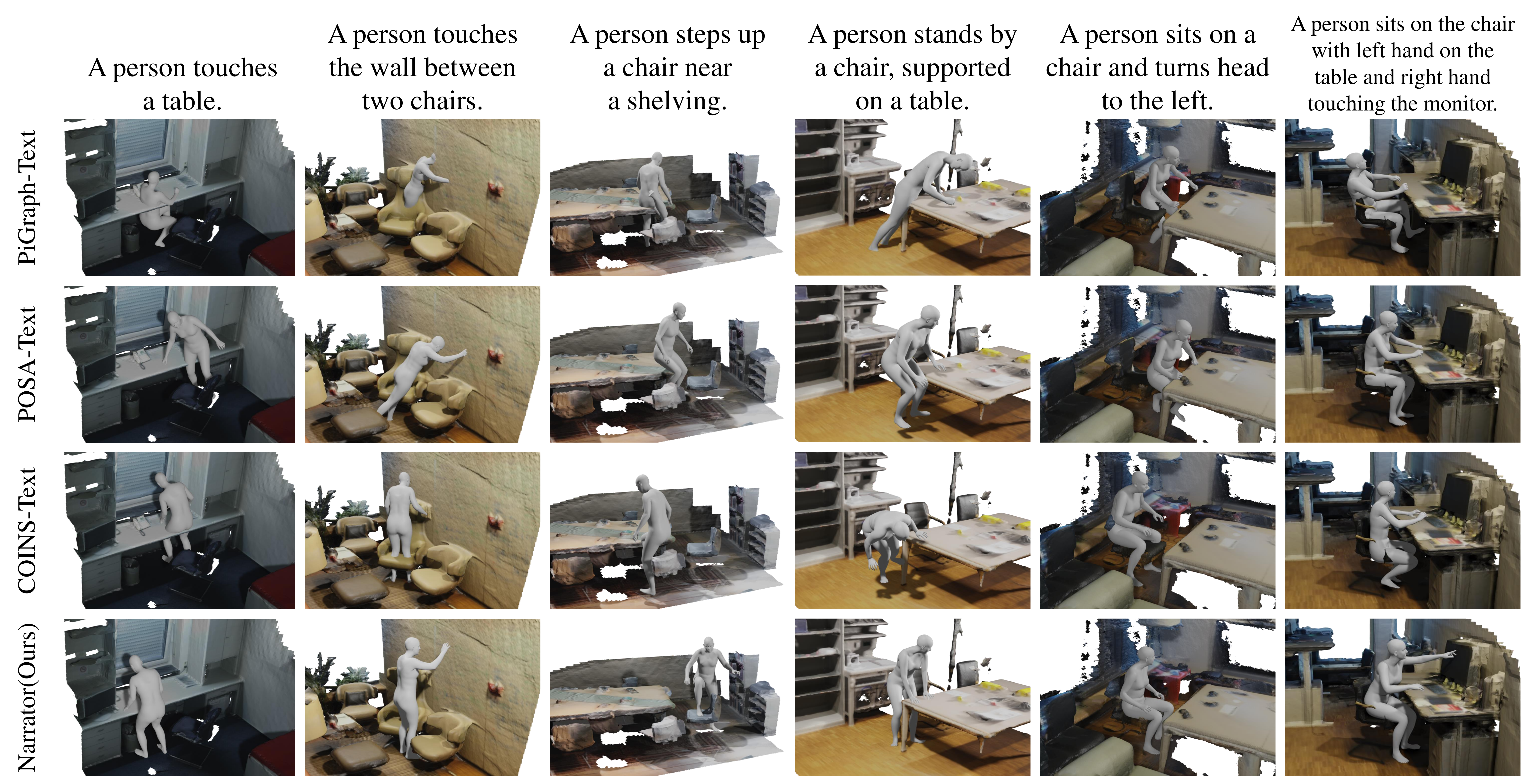}
  \vspace{-7mm}
  \caption{Qualitative comparison of interactions generated with our approach and three baselines. We present different textual queries in columns and different methods in rows. Overall, our interaction generations are semantically more consistent with textual descriptions and physically more realistic with scene interactions.}
  \vspace{-4mm}
  \label{fig:results2}
\end{figure*}

\vspace{-2mm}
\subsection{Comparison}
\vspace{-1mm}
\noindent\textbf{Perceptual study.}
Fig. \ref{fig:results1} shows the results of the perception study in terms of accuracy (correspondence to the textual descriptions) and realism. 
Respondents perceive our generated results as better matching the textual descriptions compared to the three baselines, while our generations are also clearly preferred in terms of realism. 

\noindent\textbf{Quantitative comparison.} 
Tab. \ref{tab:DB} shows the quantitative results compared to the three baselines. It can be seen that our approach has the highest accuracy and the best match to the textual description. 
our approach achieves higher contact and non-collision scores in terms of physical plausibility, demonstrating our ability to to effectively alleviate scene-body interpenetration and maintain plausible contact relationships. 
As for the diversity metrics, our approach has greater cluster entropy and smaller cluster size, achieving diversity with guaranteed accuracy. 

\noindent\textbf{Qualitative comparison.} 
We further provide qualitative comparisons in Fig. \ref{fig:results2} with the three baselines. 
PiGraph-Text suffers from more severe penetration problems due to the limitations of its own representation.
POSA-Text requires finding body placements and tends to fall into local minimums during optimization, thus generating poor interactive contacts. 
COINS-Text binds actions to specific objects, making it difficult to handle complex spatial relationships. 
In contrast, our approach can produce better results by correctly reasoning about spatial relationships and profiling the human body under multiple actions, from different levels of textual descriptions. More results in the Supp.Mat.

\begin{figure*}
  \centering
  \vspace{-3mm}
  \includegraphics[width=1.0\linewidth]{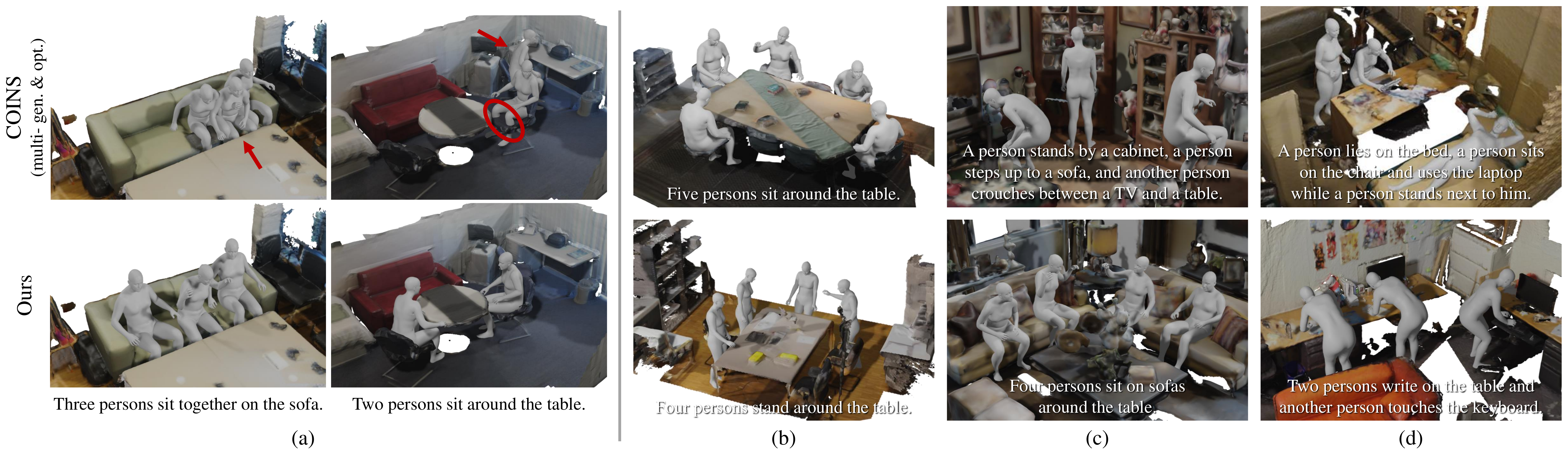}
  \vspace{-8mm}
  \caption{Qualitative comparisons with the per-person generation and optimization method using COINS \cite{zhao2022compositional} (a), and our more generation results for MHSI on PROX \cite{hassan2019resolving} (b), Matterport3D \cite{chang2017matterport3d} (c), and ScanNet \cite{dai2017scannet} (d) datasets.}
  \vspace{-4mm}
  \label{fig:results3}
\end{figure*}

\vspace{-1mm}
\subsection{Multi-Human Scene Interaction}
\vspace{-1mm}

Fig. \ref{fig:results3} (a) shows a comparison with the per-person generation and optimization method using COINS, where human collision loss is also introduced for a fair comparison. Our generation results are more natural, physically plausible, and semantically consistent, while the results of COINS are often crowded together or in the wrong position.
Additionally, our more MHSI results in different scenes are shown in Fig. \ref{fig:results3} (b), (c), and (d), including multi-person constraint, per-person constraint, and complex combination of spatial relationships and interactive actions. 
We also conduct a perception study on this and receive approval from the respondents: 84.69\% of the respondents think that the results match textual descriptions, while 74.80\% think that interactions are human-like and natural.

\vspace{-1mm}
\subsection{Ablation Study}
\begin{figure}
  \centering
  \includegraphics[width=0.9\linewidth]{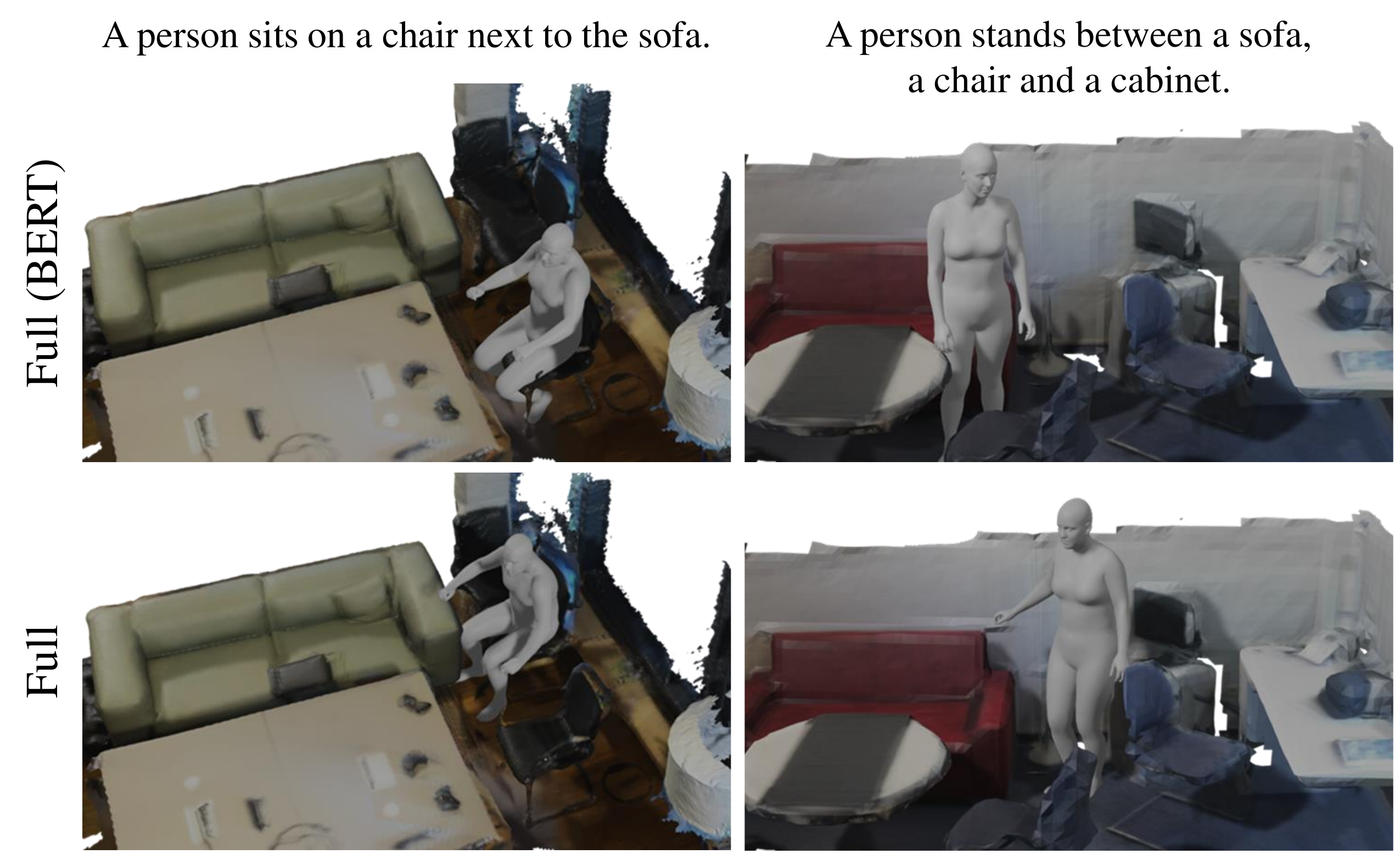}
  \vspace{-3mm}
  \caption{Qualitative results of ablation study on JGLSG. }
  \vspace{-3mm}
  \label{fig:results6}
\end{figure}

\begin{figure}
  \centering
  \includegraphics[width=0.9\linewidth]{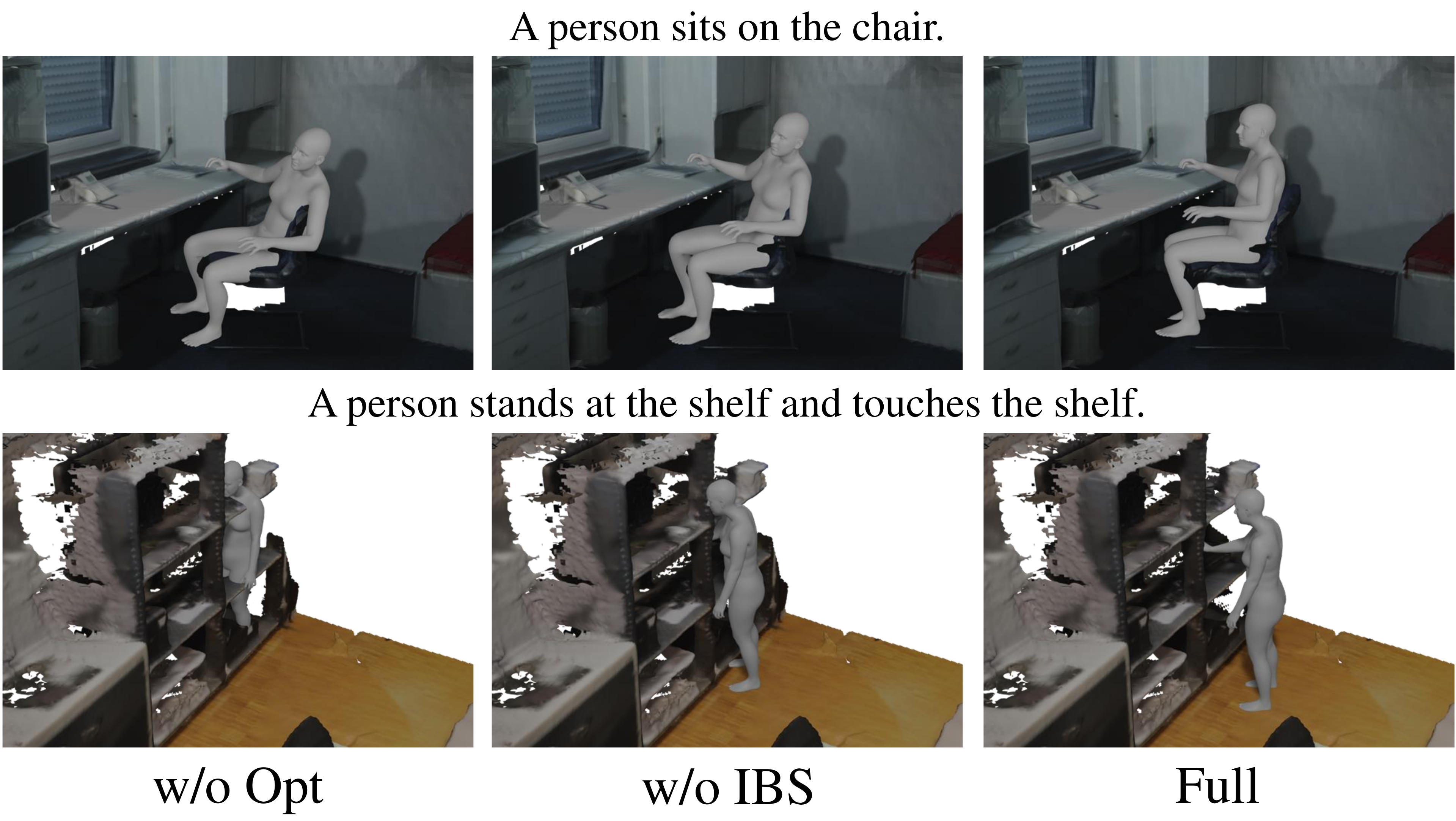}
  \vspace{-3mm}
  \caption{Qualitative results of ablation study on $\mathcal{L}_\text{IBS}$.}
  \vspace{-6mm}
  \label{fig:results7}
\end{figure}

\vspace{-1mm}
In this section, we evaluate the effect of three components of our framework. 
(1) \textbf{JGLSG.}
We study the effect of the JGLSG mechanism on interaction generation by replacing it with BERT, abbreviated as Full(BERT). 
From the qualitative comparison in Fig. \ref{fig:results6}, we can see that the interaction results obtained with the help of the JGLSG mechanism for reasoning about spatial relationships, are more accurate and more consistent with the textual description.
Quantitative results in terms of physical plausibility and diversity are shown in Tab. \ref{tab:ablation}, and again demonstrate that our approach is more effective. 
(2) \textbf{PLA.}
Tab. \ref{tab:ablation} also shows the impact of the PLA mechanism on interaction generation. Our model can refine various types of interactive actions to produce more reasonable results that are consistent with textual veracity. 
(3) \textbf{IBS loss.} 
To better handle penetration and contact issues, we additionally introduce $\mathcal{L}_\text{IBS}$. Tab. \ref{tab:ablation} and Fig. \ref{fig:results7} show comparative results using $\mathcal{L}_\text{IBS}$ and without it, demonstrating that it can improve interaction plausibility.

\begin{table}
  \centering
  \setlength{\tabcolsep}{0.7mm}
  \begin{tabular}{@{}ccccc@{}}
    \toprule
    \multirow{2}{*}{Methods} & \multicolumn{2}{c}{Physical Plausibility} & \multicolumn{2}{c}{Diversity} \\ \cmidrule(l){2-5}
    & Contact & Non-Collision & Entropy & Cluster Size \\ \midrule
    Full(BERT) & 0.92 & 0.95 & 3.82 & 1.91 \\
    w/o PLA & 0.91 & 0.89 & 3.73 & 1.66 \\
    w/o IBS & 0.90 & 0.94 & 3.95 & 1.87 \\
    Full & \textbf{0.94} & \textbf{0.98} & \textbf{4.16} & \textbf{1.54} \\
    \bottomrule
  \end{tabular}
  \vspace{-2.5mm}
  \caption{Quantitative results of ablation study.}
  \vspace{-5mm}
  \label{tab:ablation}
\end{table}

\vspace{-1mm}
\section{Conclusion and Discussion}
\vspace{-1mm}
\noindent\textbf{Conclusion.} In this paper, we observe from a narrator's perspective that humans describe human interactions in scenes through spatial perception and action recognition. 
We propose, \textit{Narrator}, a novel relationship reasoning-based generative approach for naturally controllable generation of human scene interactions from textual descriptions.
We design a JGLSG mechanism to reason about spatial relationships, and introduce a PLA mechanism for diverse interactive actions. 
In particular, benefiting from relationship reasoning, we further propose the first naturally controllable generation strategy for multi-human scene interaction. 
Experimental results demonstrate that our approach can controllably generate complex and diverse interactions and significantly outperform existing works.

\noindent\textbf{Limitation.}
Our work is mainly limited to two aspects: (1) there is still room for expansion in our interaction categories, which requires large-scale datasets; (2) our approach focuses on static interactions, while dynamic interactions are also interesting and meaningful. In further work, we will explore richer datasets and extend the scene graph to address more diverse situations, such as those involving the properties of objects or human movements.

{\small
\bibliographystyle{ieee_fullname}
\bibliography{ref}
}

\end{document}